\documentclass[12pt]{article}

\usepackage{times}
\usepackage{graphicx}

\newenvironment{sciabstract}{%
\begin{quote} \bf}
{\end{quote}}

\newcounter{lastnote}

\title{Challenges ahead Electron Microscopy for Structural Biology from the Image Processing point of view} 

\author
{Carlos Oscar S. Sorzano,$^{1\ast,2}$ and Jose Mar\'ia Carazo$^{1,2}$\\
\\
\normalsize{$^{1}$Dept. Macromolecular Structure, National Center of Biotechnology (CSIC)}\\
\normalsize{c/Darwin, 3, Univ. Aut\'onoma de Madrid, Cantoblanco, 28049 Madrid, Spain}\\
\normalsize{$^{2}$INSTRUCT Image Processing Center}\\
\\
\normalsize{$^\ast$To whom correspondence should be addressed; E-mail:  coss@cnb.csic.es.}
}
\date{}

\begin{document} 
\baselineskip24pt
\maketitle 
\includegraphics[width=14cm]{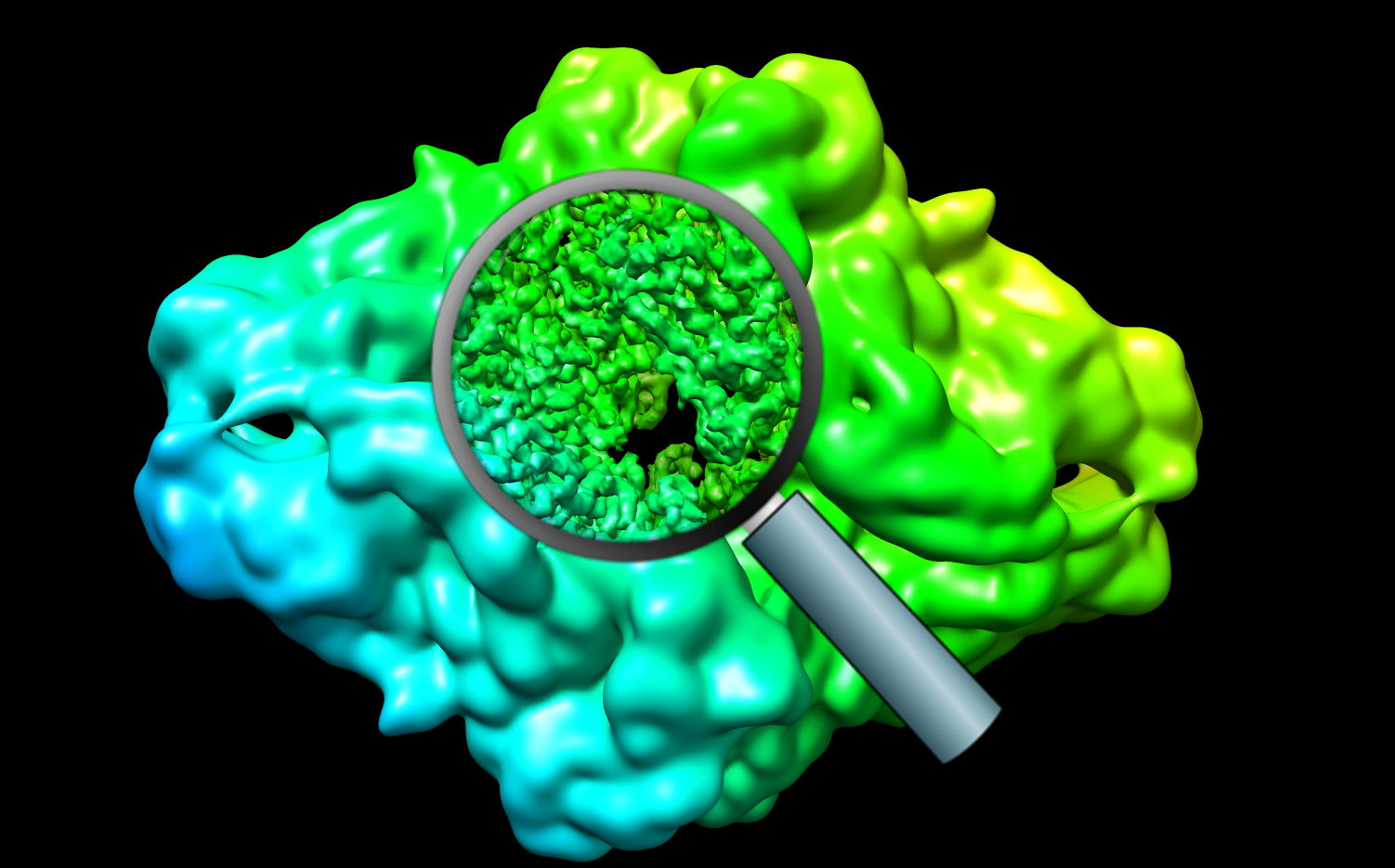}

\begin{sciabstract}
  Since the introduction of Direct Electron Detectors (DEDs), the resolution and range of macromolecules amenable to this technique has significantly widened, generating a broad interest that explains the well over a dozen reviews in top journal in the last two years. Similarly, the number of job offers to lead EM groups and/or coordinate EM facilities has exploded, and FEI (the main microscope manufacturer for Life Sciences) has received more than 100 orders of high-end electron microscopes by summer 2016. Strategic corporate movements are also happening, with very big players entering the market through key acquisitions (Thermo Fisher has recently bought FEI for \$4.2B), partly attracted by new Pharma interest in the field, now perceived to be in a position to impact structure-based drug design. The scientific perspectives are indeed extremely positive but, in these moments of well-founded generalized optimists, we want to make a reflection on some of the hurdles ahead us, since they certainly exist and they indeed limit the informational content of cryoEM projects. Here we focus on image processing aspects, particularly in the so-called area of Single Particle Analysis,  discussing some of the current resolution and high-throughput limiting factors.
\end{sciabstract}

\section*{Perspective}

In the last 2 or 3 years Electron Microscopy (EM) has undergone a profound revolution in the resolution and particle size amenable to its analysis \cite{Kuehlbrandt2014, Smith2014a}. The big promise is that it will produce high resolution structures of large and small macromolecules, as well as their dynamics in almost a native environment \cite{Nogales2016, Subramaniam2016}. Truly, the current ``world resolution record'' is 1.8\AA \cite{Merk2016} for a particle of 334 kDa and a resolution of 3.8\AA~has been obtained for a structure with weight below 100kDa (two technological barriers that were thought to prevent EM from coming useful for structure-based drug design). As a drug progresses in its design pipeline, the running costs grow by more than 7 orders of magnitude \cite{Young2009} from its beginning to the end. For this reason, having a faithful atomic model of the target allows better predictions of the binding properties of a given compound and maximize the success probability of new drugs development.

Not without reason there is an euphoria in the Electron Microscopy community about the possibilities of the field ahead. This success can be explained by a collaborative world-wide effort to improve every possible aspect of the whole pipeline: sample preparation, instrument thermal stability, electron optics improvements, image acquisition, data analysis and computational performance. Most notably, the element that has allowed the field to give a quantum leap is the recent introduction of Direct Electron Detectors (DEDs). Before DEDs were introduced, the most common way to record images was with CCDs or photographic films, which required electrons to be converted into photons by some transducer element, resulting in a severe loss of resolution and detection efficiency; additionally, these new DEDs have a very fast acquisition rate, so that usually "videos", instead of still images, are recorded. These videos have allowed to realize that the sample was moving under the electron beam during the acquisition (the so called BIM, for Beam Induced Movement), further resulting into image blurring. 

There are already relevant voices within the field that compel not to ``rest on the laurels'' and keep on working on those aspects that currently limit the routine achievement of high-resolution structures (when we remove the largest resolution problem, and certainly the old cameras was a problem of this kind, the second problem in the list becomes the first one) \cite{Glaeser2016, Vinothkumar2016}. Indeed, the list of areas to be improved is large, including sample preparation \cite{Glaeser2016}, camera detection efficiency \cite{Glaeser2016, Nogales2016, Vinothkumar2016}, specimen stabilization under the beam \cite{Vinothkumar2016}, better electron optics \cite{Zhang2011, Yang2012c, Schroeder2015b} (energy filters, aberration corrections), in-focus phase contrast \cite{Glaeser2016}, computational means for structure validity checks \cite{Henderson2012, Glaeser2016, Nogales2016}, wider access to high-end microscopes \cite{Glaeser2016, Vinothkumar2016}, and better training \cite{Glaeser2016}.

From the data analysis point of view, we would like to complement this list with some other issues. For the sake of clarity and order, we address them following the order of a typical image processing sequence:
\begin{enumerate}
	\item \underline{Better use of cryoEM video data}: Here we are witnessing from new ways to prepare EM grids \cite{Passmore2016,Thompson2016}, to algorithms for frame alignment \cite{Abrishami2015} and dose weighting \cite{Grant2015, Spear2015}, all aimed at reducing the effect of BIM. The steady progress in this area is clear and very positive. Still, the best way to  combine all these approaches is laborious and with some subjectivity. Additionally, some BIM effects are not yet addressed by any method, such us a 3D motion correction, considering out-of-plane movements (rocking). 
	
	\item \underline{Finer aberration corrections}: Microscope aberrations that have not been corrected by hardware should be estimated and corrected by software. Many attempts have been done to correct for spherical aberrations \cite{Fischer2015}, magnification anisotropy \cite{Grant2015a}, or local defocus changes \cite{Voortman2012}, but their use is not widespread, probably indicating that still a better match into the processing workflow is required. It is very enlightening to see how very recent works \cite{Georges2016} keep calling the attention to conceptually very simple issues, such as magnification calibration, specially when subtle conformational changes are being addressed, and how several papers on a so ``simple'' matter have been published just in the last year \cite{Grant2015a, Zhao2015}. Obviously, there are less clear issues to be tackled, such as the fact that the microscope focus changes along the direction perpendicular to the EM grid plane or that the weak-phase approximation is violated, all depending on the size of the specimen and the targeted resolution \cite{Zhang2011, Vulovic2013c, Koeck2015}. Indeed, these effects impact the core of the reconstruction algorithms, breaking the assumptions needed for the Central Slice Theorem, implying that beyond a given resolution the reconstruction algorithm is not correctly handling the frequency coordinates. Even such an issue as accurate focus determination is far from trivial and free of errors for very high resolution, as a recent data processing challenge showed \cite{Marabini2015}; a number of limitations of current software were assessed, specially when astigmatism was present. Finally, the so much expected introduction of phase plates as a way to avoid defocusing \cite{Danev2016} poses additional challenges, since focus determination in these conditions is specially difficult.  
	
	
	
	\item \underline{Selection of macromolecule's views}: Obtaining a biochemically pure sample for microscopy is a hard task, but obtaining a structurally homogeneous one is not just difficult, it is actually against one of the most fruitful applications of cryoEM! Still, one of the first image processing tasks in SPA is the selection of those subareas in the images containing views of the macromolecules of interest, which is a complex task, specially if hundred of thousands -or even millions \cite{Fischer2015}- of these views are being processed. The trend now is to use automatic algorithms to select on the micrographs all those subimages that ``might'' be a particle. In the best case, these algorithms have a false positive range between 80 and 95\% \cite{Abrishami2013}, meaning that between 5 and 20\% of the data does not correspond to the particle of interest but to something else.
	
	
	Better algorithms are needed to decrease the false positive rate as well as to analyze the particle candidates and discard those that are not true particles \cite{Vargas2013a}. With the levels of noise in EM, particle pruning is still an open issue in the field.
	
	
	\item \underline{Handling flexibility and heterogeneity}: Particle flexibility and heterogeneity is at the same time the blessing and the curse of EM. On one side, flexibility helps to reveal the dynamics of the macromolecule under study. On the other side, only homogeneous sets of particles can be reconstructed to atomic resolution, meaning that computational tools are needed to distinguish the different conformations in the dataset. A great advance in this regard has been made in recent years. However, the issue is far from being settled, specially in those cases in which the conformational changes being addressed are not correctly modeled as a discrete collection of classes, but they explore a continuous of states, although with different probabilities. Indeed, the very successful Maximum Likelihood approach \cite{Scheres2007b,Scheres2012} explicitly considers that the set of macromolecular views can be classified into an unknown number of distinct classes. It is not only that the determination of the number of classes may be a tedious and subjective task, but it is that the modeling does not conform to the reality behind, creating instabilities in the classification/reconstruction process. This issue has been explored in some works \cite{Dashti2014, Jin2014}, but these new approaches are not in general use in cryoEM, again possibly indicating that they have to further evolve and mature before having a significant impact in the field. Still, this issue is crucial to the continuation of the success of cryoEM, with the scarcity of works in this area only indicating how difficult it is.
	
	A particularly challenging situation occur when studying a macromolecule of unknown structure. In this case two issues combine. On the one hand, the classification/reconstruction process always requires of an "initial solution", a first map from which the optimization proceeds. On the other hand, the very possible existence of a yet unknown macromolecular flexibility/heterogeneity suggest that one single initial map may not adequately describe the sample under study. Indeed, most image classification algorithms are designed as local optimizers that start from a reasonably good initial map. If this map is not available, algorithms may easily find nonsensical structures. This is the task of the initial volume algorithms. They must produce these initial maps to be refined. However, currently, there is no algorithm specifically designed with flexibility/heterogeneity in mind.

	Finally, the widespread use of cryoEM is currently hampered by the need of very demanding computational resources. The combination of large data sets, high levels of noise and lack of software optimization are at the bases of the current situation. This situation may start changing soon by the generalization of the use of GPUs, at least in some cases, but it remains to know the impact of these changes in SPA workflows

	\item \underline{Complement with other information sources}: With very few exceptions \cite{Velazquez2012}, current reconstruction processes do not consider any other source of information than the projection images produced by the microscope. After a 3D map is obtained, modeling -specially the modelling of very large macromolecular complexes- certainly benefits from other sources of information, such as cross-linking and mass spectroscopy \cite{Politis2014} or protein-protein interaction data \cite{Segura2016}. However, the explicit algorithmic incorporation of \textit{a priori} information about the type of signals (macromolecular maps) being handled is very much missing in the field. 
	

	\item \underline{Validation}: For the good and for the bad, the data analysis always produces a model of the macromolecular structure. Unfortunately, due to the high level of noise and the high dimensionality of the optimization process, the chances of getting trapped in a local minimum are not negligible. Two are the possible manifestations of a local minimum: 1) the overall shape of the structure is incorrect (despite the fact that its projections are compatible, to a certain degree, with the experimental images); this may happen specially when an initial map is provided, and it is incorrect -the problem is known in the field as model bias-; 2) small details of the structure are incorrect (the algorithm has overfitted noise so that its objective function is minimized). The first problem can be alleviated if similar maps are obtained when starting from several initial models. However, automatic algorithms capable of detecting this situation are still in need, although some attempts are already in-place \cite{Henderson2011, Heymann2015, Vargas2016}. The second case can be alleviated by independently processing two halves of the data (what is known in the field as the gold-standard \cite{Scheres2012b}, although this is an extreme case of what in Statistics is known as $k$-fold cross validation) or detecting the local resolution and applying a spatially varying filter. More data handling strategies are needed in this regard that do not imply, as they currently do, using only a half of the dataset at hand.

\end{enumerate}

Summarizing, EM has certainly experimented a boost in resolution, processing capabilities, funding, gaining scientific momentum and future expectations. We can rightfully speak of the maturation of the field resulting in an experimental technique that can compete in quality and complement results from X-ray diffraction and Nuclear Magnetic Resonance. Fortunately for the field, there are still many open problems to solve meaning that the future can be even brighter if we solve them, but still work needs to be done.


\bibliographystyle{Science}
\end{document}